\documentclass[preprint,12pt]{elsarticle}

\usepackage{amssymb}
\usepackage{amsmath}
\usepackage{xcolor}
\usepackage{comment}
\usepackage{multirow}
\usepackage{diagbox}
\newtheorem{problem}{Problem}

\journal{Smart Agricultural Technology}

\begin{document}

\begin{frontmatter}



\title{Developing and Integrating Trust Modeling into Multi-Objective Reinforcement Learning for Intelligent Agricultural Management}


\author[inst1]{Zhaoan Wang}

\author[inst2]{Wonseok Jang}

\author[inst2]{Bowen Ruan}

\author[inst3]{Jun Wang}

\author[inst1]{Shaoping Xiao}

\affiliation[inst1]{organization={Department of Mechanical Engineering, Iowa Technology Institute, University of Iowa},
            addressline={3131 Seamans Center}, 
            city={Iowa City},
            postcode={52242}, 
            state={Iowa},
            country={USA}}
            
\affiliation[inst2]{organization={Department of Marketing, University of Iowa},
            addressline={S252 Pappajohn Business Building}, 
            city={Iowa City},
            postcode={52242}, 
            state={Iowa},
            country={USA}}

\affiliation[inst3]{organization={Department of Chemical and Biochemical Engineering, Iowa Technology Institute, University of Iowa},
            addressline={3100 Seamans Center}, 
            city={Iowa City},
            postcode={52242}, 
            state={Iowa},
            country={USA}}

\begin{abstract}
Precision agriculture, enhanced by artificial intelligence (AI), offers promising tools like remote sensing, intelligent irrigation, fertilization management, and crop simulation to boost agricultural efficiency and sustainability. Reinforcement learning (RL), in particular, has outperformed traditional approaches in optimizing yields and managing resources. Yet, widespread AI adoption remains limited by discrepancies between algorithmic recommendations and farmers' practical experiences, local knowledge, and traditional practices. To bridge this gap, our study emphasizes Human-AI Interaction (HAII), specifically targeting transparency, usability, and trust in RL-driven farm management. We employ a well-established trust framework—consisting of ability, benevolence, and integrity—to construct a novel mathematical model quantifying farmers' confidence in AI-based fertilization strategies. Farmer surveys conducted specifically for this research highlight critical misalignments, and these insights are incorporated into our trust model, subsequently integrated into a multi-objective RL framework. Unlike previous methods, our approach directly embeds trust into policy optimization, ensuring AI-generated recommendations are technically robust, economically feasible, context-sensitive, and socially acceptable. By aligning technical performance with human-centered trust, this research provides a practical path toward broader AI adoption in agriculture.
\end{abstract}

\begin{highlights}
\item AI-driven precision agriculture boosts efficiency and sustainability.

\item Reinforcement learning surpasses traditional farming methods.

\item Human-AI interaction improves trust in AI farm management.

\item Novel trust model integrates farmers' real-world experiences.

\item Embedding trust into RL ensures practical AI recommendations.
\end{highlights}


Glossary

DQN - Deep Q-Network: A type of reinforcement learning algorithm that combines Q-learning with deep neural networks. DQNs use neural networks to approximate the Q-value function, enabling agents to learn optimal policies in high-dimensional or continuous state spaces.

HAII - Human-AI Interaction: A field of study focused on the design, understanding, and evaluation of systems where humans and artificial intelligence agents interact. It encompasses aspects such as usability, trust, transparency, and collaborative decision-making between humans and AI systems.

MORL - Multi-Objective Reinforcement Learning: A reinforcement learning approach that simultaneously optimizes multiple, but often conflicting, objectives, such as maximizing crop yield while maintaining farmer trust.

POMDP – Partially Observable Markov Decision Process: A framework for decision-making under environmental uncertainty, where an agent must act based on limited observations of the true system state.

RL - Reinforcement Learning: A machine learning paradigm in which an agent learns to make decisions by interacting with an environment, receiving rewards or penalties based on its actions. The goal is to learn a policy that maximizes cumulative reward over time.

\end{frontmatter}



\section{Introduction}

With the growing impact of climate change, farmers are struggling to maintain agricultural productivity as extreme weather events, such as heatwaves and droughts, disrupt field operations and threaten crop yields. Meanwhile, global food insecurity persists, with the Food and Agriculture Organization (FAO) reporting that nearly 828 million people suffered from hunger in 2022. To address these challenges, agriculture is increasingly adopting advanced technologies, including artificial intelligence (AI), aimed at improving efficiency and resilience. Innovations such as remote sensors for monitoring crops and soil conditions, large-scale drones for precision pest management, AI-powered systems for optimizing irrigation and fertilization, and advanced modeling and simulation tools are transforming farming practices. Together, these technologies define precision agriculture, a data-driven approach that promotes sustainable and efficient food production \cite{Zhang2022}.

Recent advancements in AI-driven agricultural management have sparked significant research interest, with several studies offering valuable insights. Wu \textit{et al.} \cite{Wu2022} found that reinforcement learning (RL) techniques could surpass conventional methods in agricultural management generation, achieving similar or even better crop yields while significantly reducing fertilizer usage, marking a key step toward more sustainable practices. Similarly, Sun \textit{et al.} \cite{Sun2017} examined the use of RL for optimizing irrigation, demonstrating its ability to conserve water without negatively affecting crop health, and highlighted the usefulness of Gym-DSSAT, a crop simulation platform, in managing agricultural resources. Furthermore, Wang \textit{et al.} \cite{Wang2024} reinforced the effectiveness of RL-based fertilization strategies in challenging environments, further showcasing the potential of AI to transform modern farming.

While our earlier research showed that RL-generated agricultural management strategies could be successfully implemented in various climatic conditions \cite{wang2024continual}, discussions with farmers have highlighted significant concerns. Although AI-driven policies (i.e., strategies) aim at maximizing economic outcomes, they often focus on theoretical optimal solutions rather than aligning with farmers' practical needs and experiences. This disconnect can undermine trust in AI-generated recommendations, resulting in their rejection or discontinuation in real-world farming practices.

The challenges mentioned earlier are closely related to the field of human-AI interaction (HAII), which examines how humans engage with and respond to AI technologies. HAII has gained increasing importance in ensuring the successful adoption of AI. The field focuses on user-centered design, prioritizing aspects such as usability, transparency, interpretability, and responsiveness to human inputs and expectations \cite{amershi2019}. Its goal is to create AI systems that are developed with a deep understanding of human behavior, values, and limitations, fostering more effective interactions between people and technology. Recent studies in HAII have shown that improving transparency and interpretability can significantly boost user trust, satisfaction, and long-term acceptance of AI-driven systems \cite{abdul2018}.

Trust models have a long-standing history, applied across a wide range of contexts, from personal relationships to intricate business dynamics. A variety of models have been developed to systematically define and measure trust in different scenarios. Early research on interpersonal trust laid the groundwork for understanding how trust develops and is maintained over time. For example, Lewicki and Bunker \cite{lewicki1996developing} introduced a sequential model of trust development, outlining stages of calculative, knowledge-based, and identification-based trust. Trust models have also been extensively used in organizational settings, where they emphasize the importance of fostering long-term cooperation and stable partnerships. One influential model in this area is Butler’s Four-Component Model of Trust \cite{Butler1991}, which identifies integrity, competence, consistency, and openness as key factors influencing trust within organizations. This framework is particularly useful for assessing long-term business relationships, such as those between suppliers and clients.

As technology advances and digitalization increases, trust models have evolved to address trust in technology-mediated environments. Sectors such as e-commerce, digital finance, and virtual collaboration platforms have particularly benefited from these modified frameworks. More recently, the growing interaction between humans and AI systems has led to a surge in developing specialized trust models. Ueno \textit{et al.} \cite{Ueno2022} conducted an extensive review that categorized trust models into three types: cognitive, affective, and dispositional. Cognitive models focus on rational evaluations of AI performance and dependability, while affective models examine the emotional and relational aspects of trust. Dispositional models, on the other hand, investigate personal characteristics that influence individuals' perceptions of trustworthiness.

In agriculture communities, farmers have pointed out specific instances where AI-generated recommendations clashed with traditional agricultural practices or failed to consider important local factors, such as soil properties, historical pest issues, and distinct microclimates \cite{carolan2020automated}. Additionally, previous surveys highlighted that many farmers view AI systems as "black boxes," lacking the transparency and interpretability necessary to build trust, which raises concerns about their reliability and potential long-term risks \cite{bronson2019looking}. These findings underscore the need to incorporate strong HAII principles into agricultural AI systems, ensuring that the gap between theoretical optimal solutions and real-world applicability is bridged. This aligns with broader research advocating for technological interventions that reflect user values, experiences, and practical limitations \cite{rose2021agriculture}. To gain a deeper understanding, a detailed survey with farmers to gather their perspectives and feedback on AI-generated agricultural strategies is needed. The survey should focus on identifying areas of conflict and exploring potential improvements in AI recommendations for farming practices.

To further address this crucial gap, it is essential to develop a trust model that specifically assesses and quantifies farmers' confidence in AI-driven management strategies. The insights gathered from farmer surveys will be pivotal in refining and updating the trust model, ensuring it accurately captures real-world perceptions and improves its applicability in practical farming scenarios. By integrating farmers' established habits and feedback, we aim to customize agricultural management solutions that align more closely with their needs and expectations, fostering greater acceptance and sustained adoption.

This research makes several key contributions, starting with the design of a comprehensive farmer survey that evaluates an expert fertilization plan alongside various AI-generated plans. The survey data analysis then informs the development of a quantitative trust model tailored to assess farmer trust in AI-driven agricultural recommendations. We adopt Mayer \textit{et al.}'s \cite{Mayer1995} well-established three-dimensional framework, which comprises ability, integrity, and benevolence, as this trust model offers a more precise and practical assessment than earlier theoretical models \cite{Butler1991, meyerson1996}. By incorporating explicit data from farmers’ real-world experiences, habits, expectations, and context-specific considerations, our trust model enhances its relevance to actual farming scenarios.

In addition, our study introduces a novel integration of the trust model into the RL optimization process through a multiple-objective setting. Unlike previous research \cite{Chen2020}, where trust was assessed only after policy development, our approach embeds trust evaluations within the RL training loop. This allows real-time feedback on trust during policy optimization, alongside traditional metrics like expected returns. By doing so, our approach ensures that AI-driven agricultural management strategies are not only economically optimal but also practically acceptable and aligned with farmers' expectations, thereby increasing acceptance, effectiveness, and long-term adoption of AI solutions in agriculture.

This paper is organized as follows. After the introduction, Section 2 presents the formulation of the partially observable Markov decision process (POMDP) and provides an overview of recurrent neural network (RNN)-based deep Q-learning, as well as multi-objective reinforcement learning (MORL). Section 3 outlines the simulation environment settings, defines the problem to be addressed, and generates various AI recommendations for fertilization based on different objectives, without incorporating trust. Section 4 designs and administers a survey for farmers to evaluate the AI-generated recommendations, analyzes the results, and develops the trust model utilized in this study. Section 5 integrates the trust model into the RL training process to further optimize the fertilization strategies and presents the simulation results. Finally, Section 6 concludes the paper with a summary of our findings, a discussion of implications, and suggestions for alternative solutions and directions for future research.

\section{Methodology}

 \subsection{POMDP} \label{sec21}

A Partially Observable Markov Decision Process (POMDP) extends the concept of Markov Decision Processes (MDPs) to accommodate decision-making under uncertainty due to partial observability of states. Unlike MDPs, where an agent can fully observe the current state, in a POMDP, the agent obtains indirect observations, thereby introducing inherent uncertainty regarding the true state of the environment. A POMDP can be represented by a tuple $\mathcal{P}=\left(S, s_0, A, T, O, \Omega, R \right)$, which includes the following components.

\begin{itemize}
\item A finite set of states: $S = \{s_1, \dots, s_n\}$.
\item An initial state: $s_0 \in S$.
\item A finite set of actions: $A = \{a_1, \dots, a_m\}$, where $A(s)$ denotes the set of actions available to the agent when in state $s$.
\item A state transition probability function: $T : S \times A \times S \rightarrow [0,1]$, where $T(s, a, s')$ represents the probability of transitioning from state $s$ to state $s'$ after taking action $a$. This function satisfies the condition $\sum_{s' \in S} T(s, a, s') = 1$, ensuring that the total probability of transitioning to any possible next state sums to one.
\item A finite set of observations: $O = \{o_1, \dots, o_q\}$, where $O(s)$ represents the set of possible observations the agent can receive when it is in state $s$.  
\item An observation probability function: $\Omega : S \times A \times O \rightarrow [0,1]$, where $\Omega(s', a, o)$ defines the probability of observing $o$ after taking action $a$ and arriving at state $s'$. This function also satisfies the condition $\sum_{o \in O} \Omega(s', a, o) = 1$, ensuring that the total probability of all possible observations at $s'$ sums to one.
\item A reward function: $R : S \times A \times S \rightarrow \mathbb{R}$, which assigns a numerical reward for transitioning from state $s$ to state $s'$ after taking action $a$. This function provides immediate feedback to guide the agent's learning and decision-making process.
\end{itemize}

The primary goal of an RL agent is to maximize the expected cumulative discounted reward (i.e., expected return or utility), which reflects the agent's objective of obtaining the highest possible long-term reward by making informed decisions at each time step \cite{Sutton1998}. Formally, this is expressed as:
\begin{equation}
U(s) = \mathbb{E}\left[\sum_{t=0}^{\infty} \gamma^t R(s_t, a_t, s_{t+1}) \Big| s_{t=0}=s \right]
\end{equation}
where $s_t$ represents the state of the environment at time $t$, while $a_t$ denotes the action chosen by the agent at that time. The function $R(s_t, a_t, s_{t+1})$ specifies the immediate reward the agent receives for transitioning from state $s_t$ to state $s_{t+1}$ as a result of action $a_t$. The agent seeks to optimize its decision-making to maximize this cumulative reward. The parameter $\gamma \in [0,1]$ is the discount factor, which is crucial for balancing immediate versus future rewards. A discount factor closer to 1 places greater emphasis on future rewards, making the agent prioritize long-term gains over short-term benefits. Conversely, a lower value of $\gamma$ results in more emphasis on immediate rewards, with less concern for the long-term future. This balancing act between immediate and future rewards is central to decision-making, as agents must weigh the trade-offs and make strategic choices.

In the context of partial observability, the agent does not have direct access to the exact state $s_t$ at each time step. Instead, it relies on observations, which may be incomplete, noisy, or ambiguous, to form beliefs about the underlying state of the system. This uncertainty in the agent’s perception of the environment significantly complicates the decision-making process, as the agent must infer the most likely state from its observations rather than directly observing the state\cite{Cai2023}. As a result, the agent’s decision-making is not solely focused on optimizing for immediate rewards; it must also account for the uncertainty inherent in its belief about the state. These beliefs, typically represented as a probability distribution over all possible states, guide the agent’s action by providing a way to quantify uncertainty and make informed decisions based on its current knowledge\cite{Cai2025}. 

To address this challenge, the agent typically maintains a belief state, which serves as a representation of its uncertainty about the system's true state. This belief state is a probability distribution over all possible states, and it evolves as the agent gathers more observations. Through a process known as belief updating, the agent refines its belief over time, which in turn influences the selection of future actions\cite{Li2021}. The evolving belief allows the agent to make more informed decisions but also introduces the challenge of managing this uncertainty over time. Consequently, the agent's ability to make optimal decisions is shaped not just by the immediate effects of its actions but by the long-term consequences, which depend on how its belief about the state evolves.

One way to handle partial observability in RL is to transform the POMDP problem into a corresponding MDP one in the belief space \cite{Li2023}. This belief space is a probability distribution over states, representing the agent’s uncertainty about the environment. In this formulation, the agent’s objective is to find an optimal policy in the belief space, rather than directly in the state space. This approach allows the agent to handle uncertainty by making decisions based on its belief about the state, rather than a precise state observation\cite{Li20241}. However, this transformation can be computationally demanding and may not always be feasible when transition and observation probabilities are unknown. 

Many model-based methods for solving partially observable problems rely on knowledge of the transition and observation probabilities, which are often difficult to obtain in real-world scenarios. Model-free methods, by contrast, do not require explicit knowledge of the environment's dynamics; instead, they learn from interactions with the environment and update their policies based on observed rewards and outcomes \cite{Li2024}. These methods are particularly useful in situations where the transition and observation probabilities are uncertain or too complex to model directly, making them a suitable approach for solving real-world POMDP problems.  

\subsection{RNN-based DQN}

Q-learning \cite{Watkins1992} is a widely-used model-free RL method that leverages Q-values, also referred to as action values or state-action values, to evaluate and guide action selection throughout the learning process. The Q-value, denoted as $Q(s,a)$, represents the expected cumulative reward an agent can obtain by taking action $a$ in state $s$ and subsequently following a given policy. Through iterative updates based on observed rewards and transitions, Q-learning enables an agent to learn an optimal policy without requiring prior knowledge of the environment dynamics. Traditional tabular Q-learning methods are effective for environments with finite and discrete state spaces, where Q-values can be explicitly stored and updated in a table. However, in complex real-world applications, such as agricultural management, where state and action spaces are often continuous or extremely large, tabular methods become impractical due to their inability to generalize across states efficiently. 

To address this limitation, function approximation techniques, such as deep Q-networks (DQNs) and other neural network-based methods, have been developed to estimate Q-values in a more scalable and efficient manner, enabling RL to be applied to complex problems. DQN \cite{Mnih2013} leverages deep neural networks (DNNs) to approximate Q-values, allowing the method to generalize across vast or continuous state-action spaces where traditional tabular approaches fail. A key feature of DN is the use of two neural networks with identical architectures: an evaluation Q-network $Q_e$ and a target Q-network $Q_t$. 

The evaluation network is responsible for generating Q-value estimates and is continuously updated during training, while the target network provides stable Q-value targets for learning. To prevent instability and divergence issues caused by highly correlated updates, the target network’s weights are periodically synchronized with those of the evaluation network rather than updated at every step. This mechanism improves convergence and helps mitigate oscillations in Q-value estimates. Additionally, DQN employs experience replay, a technique that stores past experiences in a buffer and randomly samples mini-batches for training. This process breaks the correlation between consecutive experiences, further stabilizing learning and enhancing sample efficiency. These improvements make DQN well-suited for solving high-dimensional RL problems.

In real-world applications, observations provided to an agent often contain partial information about the underlying environment state, making these scenarios better modeled as POMDPs rather than fully observable MDPs. Decision-making in POMDPs necessitates the agent to integrate information over time, relying on sequences of past observations rather than single, isolated observations to infer the hidden state of the environment. To address this challenge and effectively capture temporal dependencies in observation sequences, we introduced a Recurrent Neural Network (RNN) into the DQN architecture \cite{li2023model}.  Specifically, the gated recurrent unit (GRU) \cite{Cho2014} is adopted in this study, so the Q-network can retain and utilize relevant historical observations, enabling the agent to effectively model long-term dependencies in partially observable environments, reducing the ambiguity caused by incomplete observations\cite{Zhu2022}.  

In this framework, our DQN incorporates two GRU-based Q-networks: an evaluation network and a target network. These networks are represented as $Q_E(\mathbf{o}_t,a_t;\theta_E)$ and $Q_T(\mathbf{o}_t,a_t;\theta_T)$, respectively, where $\theta_E$ and $\theta_T$ denote their corresponding sets of trainable parameters. At each discrete time step $t$, the agent observes a partial representation of the environment, generates a sequence of observations with past information, denoted as $\mathbf{o}_t$, and selects an action $a_t$ based on the Q-values predicted by the evaluation network. The selection process follows an $\epsilon$-greedy technique, which balances exploration and exploitation: with probability $\epsilon$, the agent explores by selecting a random action, and with probability 1-$\epsilon$, it exploits by choosing the action with the highest predicted Q-value. This technique ensures that the agent continues discovering new strategies while prioritizing high-reward actions.

Once the agent executes the selected action $a_t$, the environment transitions to a new state, generating a new observation, and providing a scalar reward $r_t$ that reflects the immediate benefits of the action. The observation sequence is updated by incorporating the new observation into $\mathbf{o}_{t+1}$, and the agent stores the experience as a tuple $\left(\mathbf{o}_t,a_t,r_t,\mathbf{o}_{t+1}\right)$ in an experience replay memory \cite{Lin1992}. This memory buffer plays a crucial role in stabilizing training by allowing the evaluation network to learn from a diverse set of past experiences rather than sequentially correlated transitions, which can introduce bias and hinder convergence. Specifically, the evaluation Q-network is updated using the Bellman equation given in Equation \eqref{eq:qvalue}:
\begin{equation}
\begin{split}
Q_{\text{new}}(\mathbf{o}_t, a_t) &= Q_E(\mathbf{o}_t,a_t; \theta_E) \\
&+
\alpha \left[ r_t +\gamma \max_{a'\in A}Q_t(\mathbf{o}_{t+1},a_{t+1}; \theta_T)-Q_E(\mathbf{o}_t,a_t; \theta_E) \right]
\end{split}
\label{eq:qvalue}
\end{equation}
where $\alpha$ is the learning rate. The term $\max_{a'\in A}Q_t(\mathbf{o}_{t+1},a_{t+1}; \theta_T)$ represents the highest Q-value predicted by the target network for the update observation sequence $\mathbf{o}_{t+1}$, ensuring that the agent accounts for long-term rewards when making decisions.

During training, mini-batches of experience tuples are randomly sampled from the replay buffer to update the evaluation network. This batch-based learning strategy helps break the correlation between consecutive experiences, leading to more robust learning dynamics. Meanwhile, to maintain training stability, the target network parameters $\theta_T$ are not updated continuously but rather copied periodically from the evaluation network. This delayed update mechanism prevents oscillations in Q-value targets and reduces the risk of divergence. Through the combination of recurrent network architectures, experience replay, and separate Q-networks, our approach effectively captures temporal dependencies in sequential decision-making tasks while ensuring stable and efficient learning.

\subsection{MORL} \label{sec23}

In RL, the traditional framework typically uses a single scalar reward function, $R(s,a,s')$, to guide the agent's decision-making process. However, real-world applications often involve multiple, potentially conflicting objectives that are not easily captured by a single reward. To address this complexity, our study employs Multi-Objective Reinforcement Learning (MORL) within the framework of POMDPs. Unlike traditional RL, MORL extends the single-objective reward function to a multi-objective reward vector, $\vec{R}(s, a, s')$, which enables the agent to consider several competing objectives simultaneously \cite{vamplew2011empirical}. 

This extension is formalized as: 
\begin{equation}
\vec{R}(s,a,s') = \left(R_1(s,a,s'), R_2(s,a,s'), \dots, R_k(s,a,s')\right)
\end{equation}
where each component $R_i(s,a,s')$ corresponds to a distinct objective, such as performance, efficiency, or safety. In this framework, the agent’s goal shifts from optimizing a single scalar quantity to simultaneously optimizing a vector of objectives, each of which may have its own distinct rewards associated with different state-action transitions. This more complex reward structure reflects the real-world scenario where multiple factors must be balanced, often in the presence of trade-offs and conflicts, making more informed and adaptable decisions in dynamic environments.

Due to the multiplicity and potential conflict among objectives, optimality within MORL is defined through the concept of Pareto dominance rather than scalar maximization employed in standard RL. A policy is considered Pareto optimal if no other policy exists that can improve all objectives simultaneously without causing at least one objective to deteriorate \cite{van2014multi}. The set of all such Pareto-optimal policies forms what is known as the Pareto front, representing the spectrum of optimal trade-offs among competing objectives. 

The Pareto front serves as a critical tool for decision-making in MORL, offering a structured view of the diverse solutions available. Instead of relying on a single, predefined objective function, decision-makers select policies aligned with dynamic preferences and specific contextual constraints \cite{roijers2013survey}. This flexibility is particularly valuable in complex, real-world scenarios such as autonomous driving, medical decision-making, and robotic control, where trade-offs must be carefully managed. By leveraging the Pareto front, MORL enables adaptive and informed decision-making, ensuring that chosen policies reflect the most appropriate balance between competing goals.

In practice, there are multiple strategies for choosing a final policy from the Pareto front. One common approach is a posteriori selection: after learning, the entire set of Pareto-optimal solutions is presented to a decision-maker, who then picks the most suitable policy based on domain knowledge or situational requirements. Alternatively, if user preferences are known or can be elicited at deployment time, scalarization methods (e.g., linear weighted sums, Chebyshev metrics, or other utility-based measures) can be applied to evaluate each Pareto solution according to the current preference, and the policy with the best scalar score is selected \cite{mossalam2016multi}. This combination, preserving the Pareto structure during learning, then applying preference-based filtering or ranking at selection, retains the flexibility of the multi-objective approach, while enabling decision-makers to adaptively choose among potentially conflicting objectives.  

The extension of the Q-learning algorithm to MORL contexts involves modifying the conventional scalar Bellman update rule to incorporate vector-valued rewards and Pareto dominance. This ensures action-value estimates systematically account for multiple objectives simultaneously. The modified MORL update rule is formally expressed as:

\begin{equation}
\label{eq:MORL_update}
\hat{Q}_{\text{new}}(\mathbf{o}, a) = \hat{Q}_e(\mathbf{o}, a; \theta_e) + \alpha \left[\vec{R}(s,a,s') + \gamma P(Q(\mathbf{o}')) - \hat{Q}_e(\mathbf{o},a;\theta_e)\right]
\end{equation}
where $P(Q(\mathbf{o}'))$ represents the Pareto front for the subsequent observation sequence $\mathbf{o}'$, defined explicitly as:
\begin{equation}
\label{eq:pareto_set}
P(Q(\mathbf{o}')) = {Q(\mathbf{o}', a') \mid a' \in A}
\end{equation}

This formulation differs significantly from single-objective Q-learning, which relies on the maximum operator to propagate values based on a scalar reward signal. Instead, the MORL update rule maintains the structure of the Pareto front, ensuring that value propagation respects the multi-objective nature of the problem \cite{mossalam2016multi}. By doing so, this approach prevents the collapse of multi-dimensional rewards into a single metric, preserving the integrity of trade-offs between competing objectives \cite{liu2014multiobjective}. Consequently, this principled approach enables more nuanced and informed policy selection, crucial in contexts where multiple objectives dynamically compete. 

By explicitly tracking and maintaining the Pareto front throughout the learning process, MORL algorithms ensure that policies remain reflective of the intricate trade-offs and variability inherent to real-world decision-making environments. In scenarios where a specific preference emerges only after training, practitioners can simply apply a weighting or utility-based metric to the learned front and select the corresponding strategy—thereby avoiding costly re-training under new preferences, while still benefiting from a diverse spectrum of well-balanced solutions.

\section{Intelligent Agricultural Management}

\subsection{Agricultural environment and reinforcement learning}\label{sec31}

In this research, we utilized Gym-DSSAT \cite{Gautron2022}, an advanced virtual simulation platform tailored to model crop growth, yield performance, and environmental effects like nitrate leaching. This simulation takes into account different weather conditions and initial soil parameters, offering a reliable framework for evaluating agricultural practices and management approaches. Gym-DSSAT includes 28 internal state variables that capture essential environmental and physiological factors, such as soil moisture, climate conditions, and crop growth stages. This comprehensive set of variables facilitates precise and consistent agricultural modeling.

\begin{table}[htbp]
\caption{State variables of the agricultural environment used in this study as observations.}
\begin{center}
\begin{tabular}{ | p{2.5cm} | p{9cm} | }
\hline
 \textbf{cumsumfert} & cumulative nitrogen fertilizer applications (kg/ha) 
 \\ \hline
 \textbf{dap} & days after planting  \\
 \hline
 \textbf{istage} & DSSAT maize growing stage  \\ 
 \hline
  \textbf{pltpop} & plant population density (plant/m$^2$) \\ 
 \hline
 \textbf{rain} & rainfall for the current day (mm/d)	 \\
 \hline
 \textbf{sw} & volumetric soil water content in soil layers (cm$^3$ [water] / cm$^3$ [soil]) \\
 \hline
 \textbf{tmax} & maximum temperature for the current day ($^{\circ}$C)  \\
 \hline
 \textbf{tmin} & minimum temperature for the current day ($^{\circ}$C) \\
 \hline
 \textbf{vstage} & vegetative growth stage (number of leaves)   \\
 \hline
 \textbf{xlai} & plant population leaf area index   \\
 \hline
\end{tabular}
\label{Statevariables}
\end{center}
\end{table}

However, as highlighted in our previous work \cite{Wang2024, Wang20241}, agricultural systems are inherently complex and subject to partial observability due to the wide range of interacting factors, including variable weather patterns, soil diversity, and plant behavior. To manage the uncertainty that comes with these variables, we selected ten essential state variables as the core observational inputs for the decision-making agent. These variables, outlined in Table \ref{Statevariables}, were carefully chosen for their relevance to real-world agricultural monitoring and their ability to be measured practically, ensuring that the insights from our simulation can be effectively applied to actual farming operations.

\begin{figure}
\centering
\resizebox*{10cm}{!}{\includegraphics{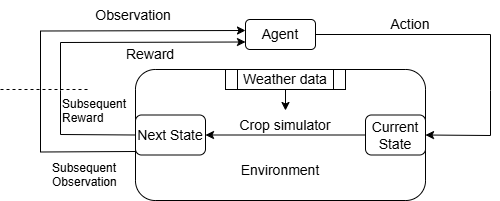}}
\caption{The interaction between an RL agent and the agricultural environment.} \label{fig:POMDP}
\end{figure}

The interaction between the RL agent and the agricultural environment, modeled using the DSSAT simulator, is illustrated in Figure \ref{fig:POMDP}. In this setup, the RL agent continuously interacts with the simulated environment to learn optimal agricultural management strategies tailored to specific weather patterns. Given that corn cultivation in Iowa predominantly relies on natural rainfall rather than artificial irrigation \cite{Wang2024}, we intentionally excluded irrigation practices from our analysis. Instead, we focused exclusively on optimizing nitrogen fertilizer application strategy, a critical factor influencing crop health, productivity, and environmental impact. Proper nitrogen management is essential for balancing crop nutrient requirements while mitigating potential issues like nitrate leaching. 

To enable the RL agent to explore and learn effective fertilizer application strategies, we defined its action space as a range of nitrogen application rates per day. These rates are discretized into increments of 5 kg/ha, ensuring a balance between granularity and computational feasibility. Mathematically, this results in a discrete action variable, 
$k$, which takes integer values from 0 to 40. Each value corresponds to a specific daily nitrogen application rate, ranging from 0 kg/ha (no nitrogen applied) to a maximum of 200 kg/ha. By systematically evaluating different nitrogen application strategies, the RL agent aims to identify policies that enhance crop yield while optimizing resource efficiency and reducing environmental impact.

During the learning process, the agent refines its decision-making by receiving environmental feedback on its actions. This iterative process enables the agent to develop optimal nitrogen management strategies that maximize crop yield, minimize resource waste, and promote sustainable farming practices. The agent's decisions are guided by a carefully designed reward function, formulated to capture the economic, agronomic, and environmental trade-offs associated with nitrogen application. The reward function is expressed in Equation \eqref{eq:AgReward}, incorporating key factors such as economic profitability, fertilizer efficiency, and environmental sustainability:
\begin{equation} 
\label{eq:AgReward}
R_t= \left\{ \begin{array}{cc} w_1 Y - w_2 N_t - w_3 L_t - w_4 N_F & \mbox{at harvest}
\\ - w_2 N_t - w_3 L_t  & \mbox{otherwise} \end{array} \right.
\end{equation}

Here $N_t$ represents the nitrogen application on each simulation day $t$, while $L_t$ (kg/ha) denotes the corresponding nitrate leaching, a critical environmental impact factor calculated by the DSSAT simulator. The reward function differentiates between daily operations and the final harvest period. During regular growth phases, the reward penalizes excessive nitrogen use and nitrate leaching. At harvest, the reward additionally accounts for crop yield $Y$ (kg/ha), reinforcing the economic benefits of improved agricultural productivity. Moreover, the term $N_F$ captures the total number of fertilizer applications required throughout the growing season, introducing an implicit cost associated with labor and operational efforts. 

The weighting factors $w_i, i=1 ... 4$ are carefully chosen to reflect real-world economic and environmental considerations. Specifically, $w_1=0.22$ represents the market price of corn per kilogram in 2023, while $w_2=1.5$ corresponds to the price of nitrogen per kilogram. The coefficient $w_3 = 15$ assigns a penalty for nitrate leaching, set as ten times $w_2$\cite{Wang2024}, emphasizing its significant environmental consequences. Finally, $w_4 = 3$ accounts for labor costs per fertilizer application, recognizing the operational burden associated with frequent nitrogen applications. 

This reward formulation ensures that the RL agent is incentivized to adopt efficient nitrogen management practices that optimize yield while mitigating excess fertilizer use and environmental degradation. By incorporating economic and ecological factors, the model fosters a balanced approach to sustainable corn production. It should be noted that $w_3$ or $w_4$ may be set to zero depending on specific objective considerations, as discussed in the subsequent section. The problem of intelligent agricultural management is therefore defined below.

\begin{problem}\label{problem1}
    A POMDP, as defined in Section~\ref{sec21}, models the agricultural nitrogen fertilizer management task, where the environmental dynamics are simulated using the Gym-DSSAT framework. The reward function, defined in Equation~\eqref{eq:AgReward}, is formulated as a weighted sum of multiple components, including economic profit (yield), nitrogen fertilizer usage, nitrate leaching penalties, and labor costs associated with fertilizer applications. These weighting factors allow for flexibility in reflecting diverse objectives; for instance, certain scenarios may prioritize yield maximization while disregarding labor cost or environmental impact. The objective is to derive an optimal policy that generates fertilization management strategies to maximize the cumulative reward, wherein the reward structure can be tailored to align with specific economic priorities or environmental considerations.  
\end{problem}

\subsection{AI-generated recommendations} \label{sec32}

\begin{figure}[htbp]
\centering
\includegraphics[width=6.0cm]{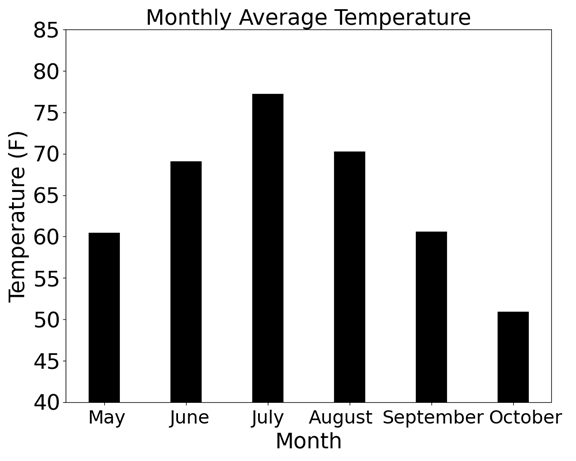}
\includegraphics[width=6.0cm]{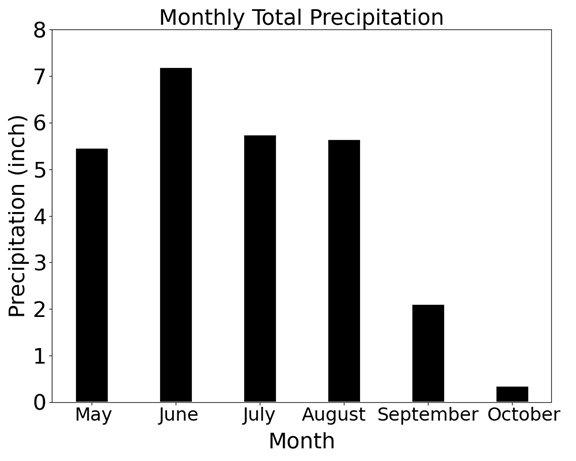}
\caption{Average monthly temperature and total precipitation during the corn growth period in Ames, Iowa, in 1999.}
\label{fig:weather}
\end{figure}

In this study, AI-generated fertilization recommendations are derived from optimal policies obtained through reinforcement learning, as described in Section~\ref{sec31}. The selected study site is a field located at the Agronomy and Agricultural Farm of Iowa State University in Ames, Iowa (42.020° N, 93.750° W). To ensure realistic and representative outcomes, we used historical weather data from Iowa in 1999, a year without extreme weather events. The average monthly temperature and total monthly precipitation during the corn growth period are illustrated in Figure~\ref{fig:weather}, providing essential environmental context for the simulation. Both the soil characteristics and historical weather data were obtained from DSSAT.

We consider four distinct scenarios, each reflecting unique management practices and environmental considerations. These scenarios target specific economic and agronomic objectives by adjusting the relative importance of the reward function components through weighting factors $w_i, i=1 ... 4$ as defined in Equation \eqref{eq:AgReward}.
\begin{itemize}
    \item Scenario 1 prioritizes environmental outcomes by significantly reducing nitrate leaching, while disregarding labor costs, setting $w_4=0$.
    \item Scenario 2 omits both nitrate leaching and labor cost considerations, setting $w_3 = w_4 = 0$. 
    \item Scenario 3 considers both nitrate leaching and labor costs, applying the full weighting scheme as described in Equation~\eqref{eq:AgReward}.
    \item Scenario 4 focuses exclusively on minimizing labor cost, neglecting environmental impacts by setting $w_3=0$.
\end{itemize}

\begin{figure} [htbp]
\centering
\includegraphics[width=7cm]{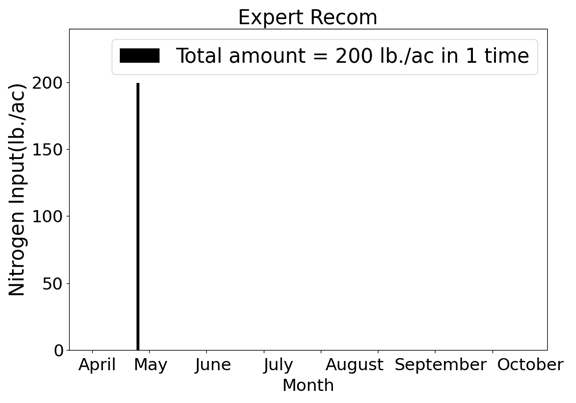}
\resizebox*{14cm}{!}{\includegraphics{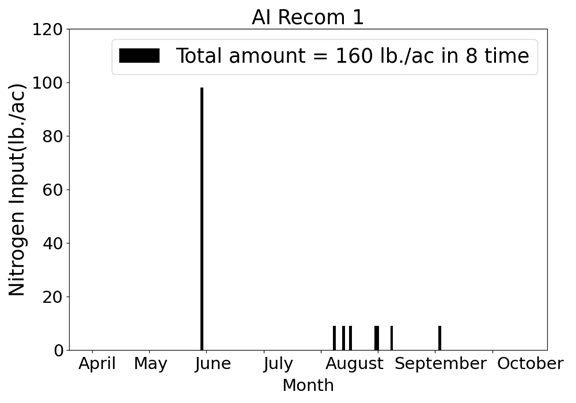},\includegraphics{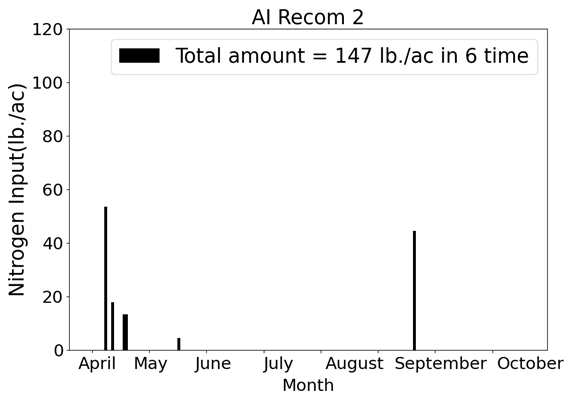}}
\resizebox*{14cm}{!}{\includegraphics{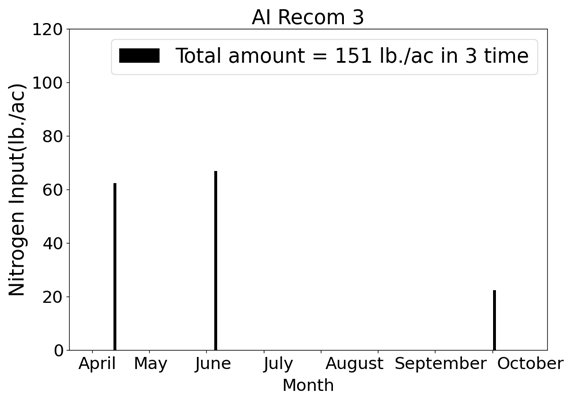},\includegraphics{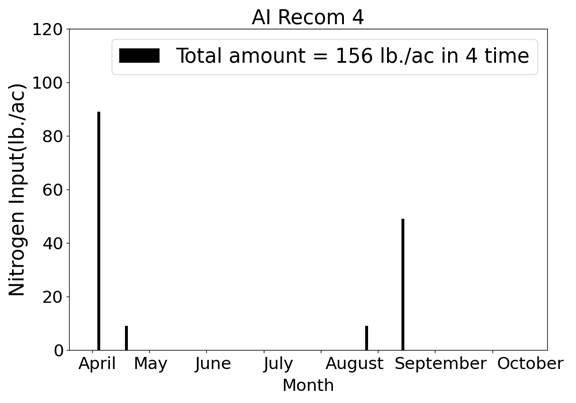}}
\caption{Five fertilization recommendations generated under different policy scenarios.}
\label{fig:policies}
\end{figure}

To learn the optimal policy for each scenario, we employ the $\varepsilon$-greedy selection strategy to balance exploration and exploitation in RL. A discount factor of 0.99 is used to emphasize the importance of future rewards in the decision-making process. The design and update of neural networks, as Q-networks, are implemented using PyTorch, with the Adam optimizer configured with an initial learning rate of $3\times10^{-5}$ and a batch size of 640. These hyperparameters are selected to strike a balance between model performance and computational efficiency. All simulations are performed on a system equipped with an Intel Core i7-12700K processor, an NVIDIA GeForce RTX 3070 Ti graphics card, and 64GB of RAM.

Once the optimal policies are learned, the corresponding fertilization recommendations - designated AI Recom 1, AI Recom 2, AI Recom 3, and AI Recom 4 - are derived and illustrated in Figure \ref{fig:policies}. For benchmarking purposes, we also include an expert recommendation (Exp Recom) from Gym-DSSAT in the figure. Each plot in Figure \ref{fig:policies} shows the timing and amount of nitrogen fertilizer applications, along with the total quantity applied and the number of application events. 

Additionally, the agricultural outcomes associated with all five fertilization recommendations are summarized in Table~\ref{table:5Recoms}. The net income, while calculated slightly differently from the reward function defined in Equation~\eqref{eq:AgReward}, is determined by subtracting the costs of fertilizer and labor from the market value of the crop yield, without accounting for environmental impacts. The expert recommendation serves as the baseline, with 100\% environmental impact corresponding to the resulting nitrate leaching. Environmental impacts from other recommendations are then quantified relative to this baseline, based on their respective levels of nitrate leaching. As a result, the AI-generated recommendations demonstrate significant reductions in environmental impact, most notably AI Recom 1 and AI Recom 3, with impacts of only 35\% and 38\%, respectively. 

Furthermore, while all recommendations achieve similar yield, AI Recom 1 results in the highest, albeit at the expense of increased labor costs due to eight fertilizer applications. AI Recom 3, meanwhile, offers an optimal balance among yield, labor cost, fertilizer usage, and environmental sustainability. It achieves slightly less yield than the expert recommendation while incurring significantly lower environmental impact and labor costs. This balance is reflected in its highest net income of \$708 per acre.

\begin{table}
\centering
\begin{tabular} {| p{2.5cm} | p{1.2cm} | p{1.5cm} | p{1.5cm} | p{0.9cm} | p{1.7cm} |p{1.3cm} | }
 \hline
   & Corn yield (bu/ac) & Number of fertilization & Fertilizer amount (lb/ac) & labor cost (\$) & Impact to environment & Net income (\$/ac) \\ [0.5ex]  
 \hline
 Exp Recom & 147 & 1 & 200 & 3 & 100\% & 684\\
 \hline
 AI Recom 1 & 147.46 & 8 & 160 & 24 & 35\% & 693  \\
 \hline
 AI Recom 2 & 146.15 & 6 & 147 & 18 & 62\% & 700  \\ 
 \hline 
 AI Recom 3 & 146.42& 3 & 151 & 9 & 38\% & 708 \\ 
 \hline
 AI Recom 4 & 147.39 & 4 & 156 & 12 & 54\% & 707 \\ 
 \hline
 \end{tabular}

\caption{Outcomes of each fertilization management recommendation.  }\label{table:5Recoms}
\end{table}

\section{Farmer Survey and Trust Model}

\subsection{Farmer Survey}

In this study, we aim to understand the factors that influence farmers' preferences and trust in agricultural management strategies. We collected data through a survey and use the insights gained to develop a novel trust model. The survey was conducted in the Midwest region of the United States. Its specific objective is to gather feedback on various fertilization management strategies recommended by either human experts or AI systems. The survey assumes normal weather conditions, with the average monthly temperature and total monthly precipitation presented in Figure \ref{fig:weather}. The focus is on nitrogen fertilizer management, particularly the timing and quantity of applications throughout the corn growth cycle. 

First, we present participants with five fertilization recommendations illustrated in Figure~\ref{fig:policies}. Participants are initially asked to rank these five recommendations on a scale from 1 (least preferred) to 5 (most preferred) without knowing whether their origin - whether they were generated by a human expert or an AI system. However, each recommendation includes information on fertilization timing and total fertilizer amount, as shown in the figure. In addition to ranking preference, participants also rate their trust in each recommendation on a scale from 1 (extremely unlikely to trust) to 5 (extremely likely to trust).

Next, we provide participants with detailed explanations of all five recommendations, as outlined in Section~\ref{sec32}, along with their corresponding outcomes listed in Table~\ref{table:5Recoms}. Specifically, we clarify that nitrate leaching from the expert-generated recommendation serves as the baseline for measuring the environmental impact of the AI-generated recommendations. Net income is calculated based on corn revenue, fertilizer costs, and labor expenses. In addition to maximizing corn yield and minimize fertilizer use, the AI-generated recommendations are optimized according to varying trade-offs between nitrate leaching — which affects environmental impact — and labor costs. After reviewing these detailed explanations, participants re-rank their preferences for the recommendations and reassess their trust ratings.

Following this reassessment, the survey asks participants to assign weights to four key decision-making factors - corn yield, fertilizer amount, fertilization frequency, and environmental impacts - ensuring that the total weight sums to 100\%. Finally,  participants respond to two questions related to their farming practices: "For seasonal
corn management, How many nitrogen fertilization applications do you believe are most appropriate?" and "How much nitrogen fertilizer do you typically apply per acre before the corn is harvested?"

\subsection{Survey data and analysis} \label{sec42}

A total of 71 farmers participated in the survey, with Iowa having the highest proportion of respondents. After excluding 4 incomplete surveys and 1 from outside the U.S., 66 complete surveys remained. Further quality checks removed problematic responses — such as duplicate rankings or uniform answers - resulting in 54 usable surveys. The final sample comprised 33 males, 19 females, 1 non-binary respondent, and 1 who preferred not to disclose their gender, with an average age of 36.9 years.

Tables~\ref{table:preferenceBefore} and \ref{table:preferenceAfter} present participants' preference ranking before and after they were informed about the source of each recommendations - whether it was expert- or AI-generated - as well as the associated outcomes. The overall ranking was calculated using a weighted average, where the percentage of responses at each ranking level served as the weight. For example, prior to receiving the explanation, 20.4\% of respondents ranked the expert recommendation as the most preferred, while 11.1\% ranked it as the least preferred, with the remaining rankings distributed as shown in Table~\ref{table:preferenceBefore}. The overall ranking score was computed as $0.204*5 + 0.185*4 + 0.259*3 + 0.241*2 + 0.111*1=3.13$. 

Initially, AI Recommendation 4 received the highest preference score (3.21), followed closely by the expert recommendation (3.13), AI Recommendation 1 (3.11) and AI Recommendation 3 (3.09). AI recommendation 2 was the least preferred, with a notably lower score of 2.47. After participants were provided with explanations, preferences shifted significantly. AI Recommendation 4 retained its top rank, but its preference score (3.70) became substatially higher than those of the other options. It was followed by AI Recommendation 2, the expert recommendation, AI Recommendation 3, and finally AI Recommendation 1, as shown in Table \ref{table:preferenceAfter}.

These findings suggest that farmers initially favored AI Recommendation 4 and the expert recommendation due to their familiarity and alignment with established farming practices. After receiving detailed explanations, the strong preference of AI Recommendation 4 persisted, indicating broad recognition of its practical advantages. In contrast, AI Recommendations 1 and 3, both of which explicitly incorporated environmental considerations, were less favorably received.

\begin{table}
\centering
\begin{tabular} {| p{2.5cm} | p{1.2cm} | p{1.2cm} | p{1.2cm} | p{1.2cm} | p{1.2cm} | p{1.3cm} |}
\hline
   & 1 & 2 & 3 & 4 & 5 & overall \\ [0.5ex] 

 \hline
Exp Recom & 11.1\% & 24.1\% & 25.9\% & 18.5\% & 20.4\% & 3.13 \\
\hline
AI Recom 1 & 20.4\% & 9.3\% & 29.6\% & 20.4\% & 20.4\% & 3.11 \\
\hline
AI Recom 2 & 31.5\% & 25.9\% & 20.4\% & 9.3\% & 13.0\% & 2.47 \\ 
\hline
AI Recom 3 & 9.3\% & 33.3\% & 11.1\% & 31.5\% & 14.8\% & 3.09 \\ 
\hline
AI Recom 4 & 27.8\% & 7.4\% & 13.0\% & 20.4\% & 31.5\% & 3.21 \\ 
\hline
\end{tabular} 

\caption{Participants' preferences for each recommendation before receiving an explanation, rated on a scale from 1 (least preferred) to 5 (most preferred). }
\label{table:preferenceBefore}
\end{table}

\begin{table}
\centering
\begin{tabular} {| p{2.5cm} | p{1.2cm} | p{1.2cm} | p{1.2cm} | p{1.2cm} | p{1.2cm} | p{1.3cm} |}
 \hline
   & 1 & 2 & 3 & 4 & 5 & overall \\ [0.5ex] 
 
 \hline
 Exp Recom & 16.7\% & 37.0\% & 9.3\% & 13.0\% & 24.1\% & 2.91 \\
 \hline
  AI Recom 1 & 31.5\% & 20.4\% &16.7\% & 20.4\% & 11.1\%  & 2.60 \\
 \hline
  AI Recom 2 & 20.4\% & 11.1\% & 35.2\% & 18.5\% & 14.8\% & 2.96 \\ 
 \hline
  AI Recom 3 & 18.5\% & 20.4\% & 25.9\% & 29.6\% & 5.6\% & 2.83 \\ 
 \hline
  AI Recom 4 & 13.0\% & 11.1\% & 13.0\% & 18.5\% & 44.4\% & 3.70 \\ 
 \hline
 \end{tabular} 

\caption{Participants' preferences for each recommendation after receiving an explanation.}
\label{table:preferenceAfter}
\end{table}

Tables \ref{table:trustbefore} and \ref{table:trustafter} illustrate participants' trust in each recommendation before and after receiving detailed explanations. Higher scores in these tables indicate greater trust. Initially, the expert recommendation received a moderate trust score of 3.42, slightly trailing AI Recommendation 2, which had the highest initial trust level at 3.52. AI Recommendations 3, 4, and 1 followed with lower trust scores of 3.30, 3.21, and 3.04, respectively. However, after the explanations were provided, the expert recommendation experienced the largest increase in trust, reaching the highest trust score of 3.67. Trust in AI Recommendations 3 and 4 also rose substantially, to 3.58 and 3.39, respectively. In contrast, AI Recommendation 1 saw only a modest increase (to 3.13), while trust in AI Recommendation 2 slightly declined to 3.39, aligning more closely with its preference ranking.

\begin{table}
\centering
\begin{tabular} {| p{2.5cm} | p{1.2cm} | p{1.2cm} | p{1.2cm} | p{1.2cm} | p{1.2cm} | p{1.3cm} |}
 \hline
   & 1 & 2 & 3 & 4 & 5 & overall \\ [0.5ex] 
 
 \hline
 Exp Recom & 5.6\% & 24.1\% & 18.5\% & 25.9\% & 25.9\% & 3.42 \\
 \hline
  AI Recom 1 & 14.8\% & 16.7\% & 31.5\% & 24.1\% & 13.0\%  & 3.04 \\
 \hline
  AI Recom 2 & 5.6\% & 13.0\% & 22.2\%  & 42.6\% & 16.7\%  & 3.52\\ 
 \hline
  AI Recom 3 & 3.7\% & 24.1\% & 27.8\% & 27.8\% & 16.7\% & 3.30 \\ 
 \hline
  AI Recom 4 & 9.3\% & 24.1\% & 20.4\% & 29.6\% & 16.7\% & 3.21 \\ 
 \hline
 \end{tabular} 

\caption{Participants' trust for each recommendation before receiving an explanation, rated on a scale from 1 (extremely unlikely to trust) to 5 (extremely likely to trust).}
\label{table:trustbefore}
\end{table}

\begin{table}
\centering
\begin{tabular} {| p{2.5cm} | p{1.2cm} | p{1.2cm} | p{1.2cm} | p{1.2cm} | p{1.2cm} | p{1.3cm} |}
 \hline
   & 1 & 2 & 3 & 4 & 5 & overall \\ [0.5ex] 
 
 \hline
 Exp Recom & 9.3\% & 13.0\% & 16.7\% & 24.1\% & 37.0\% & 3.67 \\
 \hline
  AI Recom 1 & 7.4\% & 20.4\% & 35.2\% & 25.9\% & 11.1\%  & 3.13 \\
 \hline
  AI Recom 2 & 7.4\% & 9.3\% & 40.7\% & 22.2\% & 20.4\%  & 3.39 \\ 
 \hline
  AI Recom 3 & 1.9\% &  14.8\% & 27.8\% & 35.2\% & 20.4\% & 3.58 \\ 
 \hline
  AI Recom 4 & 3.7\% & 14.8\% & 33.3\% & 35.2\% & 13.0\% & 3.39 \\ 
 \hline
 \end{tabular} 

\caption{Participants' trust for each recommendation after receiving an explanation.}
\label{table:trustafter}
\end{table}

Interestingly, participant preferences did not consistently align with trust ratings. Preferences were somewhat more flexible, whereas trust appeared more stable. The divergence between trust in a recommendation and an individual's ultimate preference can be understood by considering the distinct foundations of these two constructs. Trust is often grounded in a cognitive assessment of the recommendation's source. We deem a recommender trustworthy based on perceptions of its competence, benevolence, and integrity \cite{Mayer1995}. 

However, a gap often exists between an individual's level of trust and their expressed preferences. This gap arises partly because preferences are not fixed entities waiting to be discovered; rather, they are frequently constructed at the moment of decision and are highly malleable, shaped by context, choices framing, available options, and transient personal goals \cite{Simonson2008, Slovic1995}. For example, a trusted financial advisor might recommend a diversified, long-term growth fund (a suggestion reflecting competence and benevolence), but an individual's preference at that moment might be swayed by a news headline about a speculative tech stock (framing and context) or a sudden short-term financial need (current goals). 

Furthermore, while trust is often based on logical assessment, preferences, particularly for experiential choices like entertainment, food, or travel, are strongly affective in nature. They hinge on how an option makes us feel \cite{Zajonc1980}. Someone may trust a critic's expertise yet prefer a lighthearted comedy for emotional lift. In such cases, a cognitively trusted recommendation may fail to resonate on the affective level, creating a disconnect with momentary preference.

We also believe the small sample size contributed to this discrepancy. Individual variations, particularly strong anti-AI sentiments among a few participants, may have disproportionately influenced overall trends. Additionally, some participants exhibited a deep-rooted trust in human expertise, suggesting enduring attitudes rather than situational preferences. Moreover, the limited sample may not adequately capture the diversity within the farming community. Factors such as age, education, farming experience, and technology acceptance significantly influence perceptions and trust in decision-support tools. Underrepresentation of specific subgroups could thus skew overall trust assessments between AI and expert recommendations.

The justifications provided by participants suggest that some farmers prioritized the timing and frequency of fertilization. Many noted that their trust in a recommendation was primarily driven by its alignment with their existing fertilization schedule. For example, some mentioned that they were more likely to trust and adopt a strategy if it scheduled fertilizer applications in April or May, as this aligned with their traditional practices (e.g., “I think April or May is the best combination of rainfall and growing plants taking advantage of the nitrogen”). Furthermore, participants indicated that the most significant factor influencing changes in their preference was net income, which was only disclosed after the explanations were provided. Many participants explicitly stated that net income was their primary consideration (e.g., “Net income is what matters most to me”). 

Among the four decision-making factors presented in the survey - corn yield, fertilization amount, fertilization frequency, and environmental impact - farmers prioritized corn yield the highest at 40.24\%, followed by fertilization amount at 23.91\%. Fertilization frequency ranked third at 19.02\% while environmental impact was the least prioritized, with 16.83\% of participants mentioning it. 

Additionally, our survey included two general questions: "For seasonal corn management, How many nitrogen fertilization applications do you believe are most appropriate?" and "How much nitrogen fertilizer do you typically apply per acre before the corn is harvested?" Among all valid responses, about three-fourths of participants indicated that the ideal fertilization frequency is two or three applications per growth cycle. The most commonly cited fertilizer amounts were 150 lb/acre and 200 lb/acre, with other responses falling within this range.

\subsection{Trust model}

Trust is a multifaceted social and psychological phenomenon shaped by various factors, including social networks, personal charisma and appearance, confidence levels, perceived competence in task execution, and credibility established through past interactions and connections \cite{Cho2009}. It serves as the cornerstone of successful relationships between humans and non-human agents, e.g., AI agents. When these relationships involve dependence and risk - especially given the complexity and non-deterministic behavior of AI systems - trust becomes crucial. Misplaced or insufficient trust can result in misuse, overreliance, or outright rejection of the technology. Moreover, the successful integration of AI into workplace processes depends on workers’ confidence in its capabilities and reliability \cite{omrani2022}.

The conditions that foster trust have been extensively studied in the literature \cite{Butler1991,Strickland1958}. Mayer \textit{et al.} \cite{Mayer1995} synthesized these discussions into three core factors underpinning trust: ability, benevolence, and integrity, collectively forming the ``three-dimensional trust model." In this framework, ability refers to the skills, competencies, and expertise that enable an entity to exert influence within a specific domain, emphasizing the importance of demonstrated proficiency. Benevolence, by contrast, reflects the extent to which a trustee is perceived as genuinely committed to acting in the trustor’s best interest, free from self-serving motives, thereby underscoring the importance of altruistic intent. Integrity encapsulates the trustor’s belief that the trustee consistently adheres to principles aligned with their values, highlighting the role of moral consistency and ethical commitment. Together, these dimensions provide a comprehensive framework for understanding the dynamics of trust across diverse contexts.

In this study, we build upon the mechanisms outlined by Mayer \textit{et al.} \cite{Mayer1995} and broadly hypothesize that several key factors play a critical role in shaping farmers’ trust in AI-based agricultural management systems, with particular emphasis on their trust in AI agents. To explore this, we use the previously mentioned three-dimensional trust model as the foundation to develop a farmer-specific trust model, grounded in the survey data and subsequent analysis.

The first dimension of our model focuses on the AI agent’s ability, emphasizing its competence in achieving desirable outcomes. To quantify this, we derived Equation~(\ref{ability}) to approximate farmers’ trust in the AI agent’s ability to provide effective fertilization recommendations. While the five recommendations presented in the survey resulted in comparable corn yields, all exceeding the U.S. average of 8,649 kg/ha (137.8 bu/ac) reported by the USDA in 1999 \cite{USDA}, yield ($Y$) remains the most direct measure of agronomic success, as it directly impacts farmers' net incomes. Therefore, we adopted 8,649 kg/ha as the baseline value in Equation~(\ref{ability}).

\begin{equation}
\text{ability} = \frac{Y}{8649}*\frac{1}{cosh(0.1*(\sum_tN_t-196))}*\frac{0.5}{(\left|N_F-2.5\right|)}* \frac{O_F+1}{N_F+1}
\label{ability}
\end{equation}

Other key indicators, including fertilizer usage ($\sum_t N_t$), fertilization frequency ($N_F$), and the total number of fertilizer applications within the optimal window ($O_F$), also serve as measures of the AI agent's competence, as they reflect how well its recommendations align with standard farming practices and influence farmers' trust. According to our survey, appropriate fertilizer usage ranged from 150 to 200 lb/acre (approximately 168 to 224 kg/ha), with an average of 196 kg/ha. To capture the idea that farmers are less likely to trust recommendations that significantly deviate from this average, we applied the hyperbolic cosine function $\cosh$ in Equation~(\ref{ability}) to smoothly penalize such deviations. A scaling factor of 0.1 was introduced to ensure that small deviations would not result in disproportionately large reductions in trust. Additionally, we set a baseline fertilization frequency of 2.5 in Equation~(\ref{ability}), reflecting survey responses that indicated two to three applications per growth cycle as ideal. Lastly, the optimal fertilizer application window was defined as April to May, based on the majority of survey participants' responses. 
 
The second dimension, benevolence, reflects whether farmers perceive AI agents as beneficial to their farming practices and aligned with their interests. In this study, we evaluated benevolence by examining the total amount of nitrate leaching ($\sum_t L_t$) resulting from the AI-generated recommendations, using it as a proxy for environmental impact. While most survey participants expressed limited concern about nitrate leaching — and some even viewed efforts to reduce it negatively — they generally believed that overemphasizing environmental benefits could negatively affect their income. As shown in Table~\ref{table:preferenceAfter}, recommendations that prioritized environmental outcomes were typically less preferred. However, a subset of participants noted that effective nitrate control could enhance their trust in AI-driven fertilization management, as it contributes to long-term agriculture sustainability.  

We set the nitrate leaching baseline at 0.14 kg/ha, which corresponds to the outcome of AI Recommendation 4 - a recommendation that accounted for labor costs but not nitrate leaching. This recommendation was the most preferred, selected by 44.4\% of participants. Nitrate leaching levels below this baseline may imply that the AI agent over-prioritizes environmental concerns at the expense of farmers' practical needs, whereas levels significantly above it could indicate inefficient nitrogen fertilizer use, potentially compromising both productivity and environmental health. Based on this rationale, we formulated Equation~(\ref {benevolence}) to evaluate the AI agents' benevolence in generating fertilization recommendations.

\begin{equation}
{\text{benevolence} = e^{-\big(\frac{\sum_t L_t-0.14}{0.1}\big)^2}}
\label{benevolence}
\end{equation}

The final dimension, integrity, captures farmers’ perceptions of the transparency and fairness of AI-generated recommendations. It gauges whether farmers believe the recommended agricultural management strategies are unbiased and openly designed. In our study, we propose that integrity can be assessed based on the credibility of the AI's design team. It is assumed that the AI agent is developed by an independent research group with a strong reputation in relevant fields. The group also provides clear and detailed explanations about AI's decision-making process. In this case, farmers would likely trust the integrity of the AI-generated recommendations fully (i.e., integrity = 1). 

In summary,  we use Equation~(\ref{trust_score}) to calculate the overall trust score, which allows us to quantify trust and optimize recommendations,
\begin{equation}
\text{Trust Score} = \text{ability} * \text{benevolence} * \text{integrity}
\label{trust_score}
\end{equation}

It is important to note that while the benevolence and integrity components are constrained to values between 0 and 1.0, the ability score may slightly exceed 1 due to the yield component in Equation~(\ref{ability}). By allowing this component within the ability dimension to surpass 1.0, we reflect this weighting and emphasize yield’s dominant role in shaping farmers’ trust in AI-generated fertilization policies. This design decision aligns with findings from our survey (Section~\ref{sec42}), which indicate that farmers prioritize corn yield (40.24\%) more heavily than fertilizer amount (23.91\%), fertilization frequency (19.02\%), or environmental impact (16.83\%).

Table \ref{table:trustscore} presents the trust scores for AI-generated recommendations based on the three-dimensional trust model, Equation~(\ref{trust_score}), developed in this study. Although all scores are relatively low, reflecting the AI agent's approach to intelligent agricultural management without considering farmers' trust, the order of these scores closely matches the trust rankings reported by survey participants (Table~\ref{table:trustafter}). AI Recommendations 3 and 4 receive the highest trust scores, primarily due to their optimal fertilization schedules and well-balanced nitrate leaching levels. AI Recommendation 2 earns a moderate trust score, indicating a reasonable but less optimal alignment with farmers' expectations and perceived interests. Moreover, AI Recommendation 1, however, receives the lowest trust score, highlighting a significant disconnect with farmers' priorities, as evidenced by its ranking in Table~\ref{table:trustafter}.

\begin{table}
\centering
\begin{tabular} {| p{2.5cm} | p{2.5cm} |}
 \hline
   & Trust score \\ [0.5ex] 
 
 \hline
 AI Recom 1 & 0.0002 \\
 \hline
 AI Recom 2 & 0.01\\ 
 \hline
 AI Recom 3 & 0.04 \\ 
 \hline
 AI Recom 4 &  0.04 \\ 
 \hline
 \end{tabular} 

\caption{Trust scores for AI-generated recommendations}
\label{table:trustscore}
\end{table}

Interestingly, the expert recommendation receives a low trust score of 0.01, which contrasts with its relatively high ranking based on participants' survey responses (Table~\ref{table:trustafter}). This discrepancy can be attributed to the fact that participants' trust in the expert recommendation was likely influenced more by a general confidence in human expertise than by a detailed evaluation of the recommendation itself. In other words, while participants may have rated the expert recommendation highly due to an overarching trust in human knowledge and experience, they may not have closely examined the specific content of the recommendation when assessing its trustworthiness. 

Such behavior is consistent with findings from previous studies. Despite demonstrated advantages of AI agents, individuals often exhibit a reluctance to rely on algorithmic decision-making — a phenomenon known as algorithm aversion. This refers to the tendency to lose trust in algorithms after observing even minor errors, even when those algorithms outperform human decision-makers on average \cite{Dietvorst2015}. This pattern highlights a key distinction: whereas trust in human experts often draws from generalized, affective, or identity-based beliefs, trust in AI-generated recommendations tends to require more deliberate, evidence-based justification.

\section{Trust-aware Intelligent Agricultural Management} \label{sec5}

In this section, we aim to address the challenge of agricultural nitrogen management by jointly considering two primary, and potentially conflicting, objectives: (1) maximizing agronomic performance, which includes optimizing crop yield, fertilizer usage, and nitrate leaching control, and (2) maintaining a high level of trust in the AI agent's decision-making. These objectives can be in tension; for instance, strategies that overly prioritize nitrate control may conflict with farmers' preferences and reduce trust, while trust-enhancing approaches may compromise task efficiency. To isolate the impact of trust, we set $w_4=0$ in the reward function (Equation~(\ref{eq:AgReward})) while keeping the other parameters the same in this section. Accordingly, the problem is modeled as an MORL problem, and the statement in Problem~\ref{problem1} is revised as follows.

\begin{problem}\label{problem2}
    A POMDP, as defined in Section~\ref{sec21}, models the agricultural nitrogen fertilizer management task, where the environment dynamics are simulated using the Gym-DSSAT framework. A primary reward function (Equation~(\ref{eq:AgReward})) evaluates agronomic performance by jointly considering crop yield, fertilizer usage, and environmental impact. In this formulation, the labor cost component is excluded, allowing the focus to remain on economic return and environmental sustainability. Additionally, a trust model (Equation~(\ref{trust_score})) quantifies farmers' trust in the AI agent's recommendations. The objective is to learn an optimal policy that generates trust-aware fertilization strategies by simultaneously maximizing the cumulative reward and the trust score. 
\end{problem}

To address this problem, we adopted an MORL approach using DQN, as detailed in Section~\ref{sec23}, to simultaneously maximize agricultural rewards and farmers' trust in fertilization recommendations. MORL allowed us to systematically explore and balance these competing objectives by generating a set of Pareto-optimal solutions. Upon convergence, the MORL framework produced a Pareto front, which is a collection of non-dominated policies representing optimal trade-offs between objectives. To select a final policy from the Pareto front, we employed an explicit preference-weighting strategy, assigning equal importance (50:50) to agricultural rewards and farmers' trust. This balanced approach ensured that the selected recommendations emphasized both agronomic effectiveness and perceived trustworthiness.

\subsection{Trust-aware AI-generated recommendation}

We first re-studied the intelligent agricultural management subject to the normal weather of 1999, as shown in Figure~\ref{fig:weather}. After the optimal policy was learned by the AI agent, a fertilization recommendation could be generated. We referred this AI-generated recommendation as trust-aware because the agent considered farmers' trust during the learning process. As a comparison, we chose AI Recommendation 1 introduced in Section 3.2 since its has the same agronomic objectives: maximizing crop yield while minimizing fertilizer usage and nitrate leaching. However, the AI agent didn' consider farmers' trust when learning and geneating AI Recommendation 1, so it was referred to as trust-agnostic. Additionally, we added the expert recommendation from Gym-DSSAT into the comparison. Table~\ref{table:result} summarizes the agricultural outcomes from three different fertilization recommendations. 

\begin{table}[htbp]
\begin{center}
\begin{tabular}{| p{5cm} | p{2.4cm} | p{1.5cm} | p{2.6cm} |}
\hline
  Recommendation & Trust-aware & Expert & Trust-agnostic   
 \\ \hline
 Total reward & 1747 & 1697 & 1763  \\
 \hline
 Yield (Kg/ha) & 9245 & 9248 & 9248  \\
 \hline
 Nitrogen input (Kg/ha) & 190 & 224 & 180  \\
 \hline
 Nitrate leaching (Kg/ha) & 0.12 & 0.26 & 0.09 \\
 \hline
 Fertilization frequency & 2 & 1 & 8 \\
 \hline
 The number of nitrogen applciations in April and May & 2 & 1 & 1 \\
 \hline
Trust score & 0.867 & 0.01 & 0.0002 \\
 \hline
\end{tabular}
\caption{Agricultural outcomes from different recommendations.}
\label{table:result}
\end{center}
\end{table}

The results presented in the table reveal that the trust-agnostic AI-generated recommendation achieves the highest overall reward among the three recommendations strategies evaluated. This recommendation suggests its strong technical effectiveness in optimizing the targeted agronomic and environmental outcomes. However, despite its superior reward, this recommendation receives the lowest trust scores. The primary reason for this low level of trust is its failure to incorporate farmers' preferences, practices, and perceptions. Specifically, the recommendation favors small but frequent fertilizer applications as a strategy to minimize nitrate leaching. While environmentally beneficial, this approach significantly increases the operational burden on farmers by requiring more field visits and extended working hours. Additionally, the fertilizer application dates proposed by this strategy often deviate sharply from farmers’ traditional schedules, creating further misalignment with their expectations and routines and thereby exacerbating their reluctance to adopt the recommendation. 

In contrast, the expert recommendation takes a much more conservative and familiar approach. It suggests applying fertilizer only once at the beginning of the corn growing season, aligning more closely with traditional farming practices. As a result, this strategy achieves moderately higher trust scores in simulation studies, reflecting a greater level of comfort and acceptance among farmers. However, this approach also results in greater fertilizer use and consequently higher levels of nitrate leaching, which undermines its environmental sustainability and reduces its overall reward. Ultimately, despite being more trusted, the expert recommendation performs the worst in terms of reward among the three strategies.

The trust-aware AI-generated recommendation offers a compelling middle ground by balancing technical performance with anticipated social acceptability. This strategy achieves a reward comparable to the trust-agnostic recommendation, while also yielding a substantially higher trust score than the expert recommendation. Although we did not conduct a direct survey to assess farmer responses to this specific recommendation, the trust model used to guide its design was developed based on survey data capturing farmers’ behaviors and preferences. By leveraging this information, the trust-aware recommendation strategically calibrates both the frequency and timing of fertilizer applications to better align with typical farming schedules. It proposes application periods that coincide with times when farmers are usually active in their fields, thereby reducing perceived disruptions. This contextual alignment is intended to ease implementation and foster a sense of familiarity and control—factors associated with increased trust and the likelihood of adoption.

\subsection{Climate variability}

In our subsequent study, we investigate the impact of the trust model on fertilization strategies under conditions of climate variability. Since the trust model was developed based on survey data collected under the assumption of normal weather conditions, its applicability and robustness under changing climatic scenarios remain uncertain. Climate variability - such as altered precipitation patterns, temperature fluctuations, or extreme weather events — may influence both the agronomic effectiveness of fertilization strategies and farmers’ perceptions of risk, feasibility, and trustworthiness. Therefore, we aim to evaluate how trust-aware recommendations perform when subjected to a range of plausible weather conditions, and whether the trust model continues to provide reliable guidance in aligning technical optimization with human-centered preferences. This exploration is essential for assessing the long-term adaptability and resilience of trust-informed decision-making frameworks in agriculture.

We utilized weather data from 1999 (Figure~\ref{fig:weather}) as a baseline and introduced variations in temperature and precipitation to evaluate the performance of trust-aware optimal policies under different climate variability scenarios. Two scenarios were conducted: one involving an increase in temperature, and the other a reduction in precipitation. In the first scenario, we increased daily temperature (increments of +1°C, +2°C, and +5°C) relative to the 1999 baseline throughout the entire year, while maintaining the original precipitation levels. In the second scenario, we reduced daily precipitation (decreases of 20\%, 40\%, and 80\%), while keeping temperature patterns consistent with the 1999 baseline. Importantly, soil conditions remained unchanged from 1999 for all simulations. Furthermore, scenarios involving increased precipitation that could lead to flood-related crop damage were not considered, as such damage falls outside the predictive capabilities of the DSSAT model.

\begin{figure}
\centering
\resizebox*{8.5cm}{!}{\includegraphics{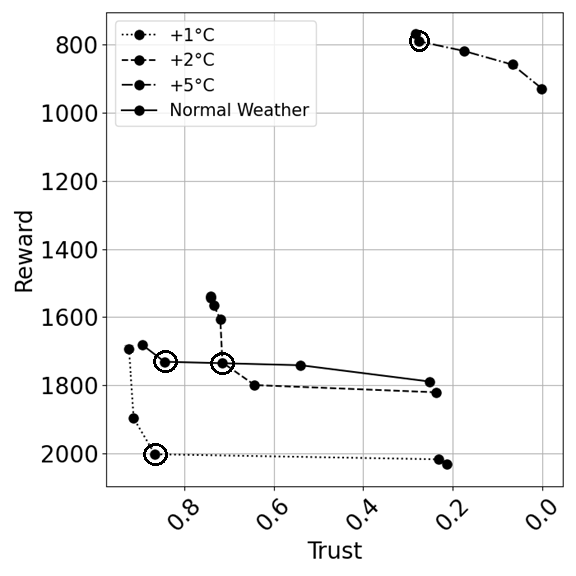}}
\caption{Pareto fronts under scenarios of of temperature increase. The circled point represents the selected policy. } \label{fig:pft}
\end{figure}

\begin{table}
\centering
\begin{tabular} {| p{5.4cm} | p{2.4cm} | p{2.6cm} | } 
 \hline
 \diagbox[innerwidth=5.4cm]{Temperature Increment}{recommendation} & Trust-aware  & Trust-agnostic  
 \\
 \hline
 \hline
  $+1^{\circ}$C &   &     
 \\ 
 \hline
 Yield (kg/ha) & 10425 & 10425  
 \\
  \hline
 Nitrogen input (kg/ha) & 190 & 160  
 \\
  \hline
 Nitrate leaching (kg/ha) & 0.10 & 0.10  
  \\
  \hline
  Fertilization frequency & 2 &  5  
 \\
  \hline
 Optimal window fertilizer count & 2 &  2 
  \\
  \hline
  Total Reward & 2007 & 2052
  \\
  \hline
  Trust Score & 0.866 & 0.003 
 \\
  \hline  \hline
   $+2^{\circ}$C &   &     
 \\ 
 \hline
 Yield (kg/ha) & 9352 & 9357  
 \\
  \hline
 Nitrogen input (kg/ha) & 190 & 120  
 \\
  \hline
 Nitrate leaching (kg/ha) & 0.09 & 0.09  
  \\
  \hline
  Fertilization frequency & 2 &  6  
 \\
  \hline
 Optimal window fertilizer count & 2 &  2 
  \\
  \hline
  Total Reward & 1771 & 1847 
  \\
  \hline
  Trust Score & 0.710 & 0.0003 
 \\
  \hline \hline 
    $+5^{\circ}$C &   &      
 \\ 
 \hline
 Yield (kg/ha) & 4901 & 4873  
 \\
  \hline
 Nitrogen input (kg/ha) & 190 &  60 
 \\
  \hline
 Nitrate leaching (kg/ha) & 0.08 &  0.07 
  \\
  \hline
  Fertilization frequency & 3 &  3  
 \\
  \hline
 Optimal window fertilizer count & 3 &  1 
  \\
  \hline
  Total Reward & 792 & 981 
  \\
  \hline
  Trust Score & 0.33 & $2.9\times10^{-7}$ 
 \\
  \hline
\end{tabular}
\caption{Comparison of different fertilization recommendations under an increased temperature scenario.}\label{table:TempImpact}
\end{table}

Figure~\ref{fig:pft} shows Pareto fronts illustrating the trade-offs between reward and trust under varying temperature increases. At each temperature level, the AI agent re-learned optimal policies using the developed MORL framework before generating the corresponding recommendations. Under mildly elevated temperatures (+1°C), trust-aware AI-generated recommendations outperform those under baseline conditions, achieving higher rewards and greater rust. This improvement likely stems from modest yield gains in crops such as corn due to slight warming, which enhances both agronomic performance and perceived reliability. At moderate warming levels (+2°C), the Pareto front still demonstrates a reasonable balance between reward and trust. Despite similar yields to the baseline, overall trust levels decline, likely due to reduced nitrate leaching, which may lower farmers' perceived justification for higher nitrogen inputs and affect confidence in the recommendations.

In contrast, under extreme heating (+5°C), both reward and trust scores deteriorate significantly, constraining the potential for meaningful trade-offs. The trust model — originally calibrated under baseline conditions (normal weather) — becomes less reliable in this regime. While trust-aware recommendations still yield marginally better outcomes than their trust-agnostic counterparts, the rationale for elevated nitrogen use weakens, as reflected in the steep decline in reward.

Table~\ref{table:TempImpact} illustrates practical implications: trust-aware recommendations consistently reduce fertilization frequency compared to trust-agnostic recommendations, albeit at slightly higher nitrogen inputs. The reduced number of applications enhances trust by minimizing labor and management burden.

\begin{figure}
\centering
\resizebox*{8.5cm}{!}{\includegraphics{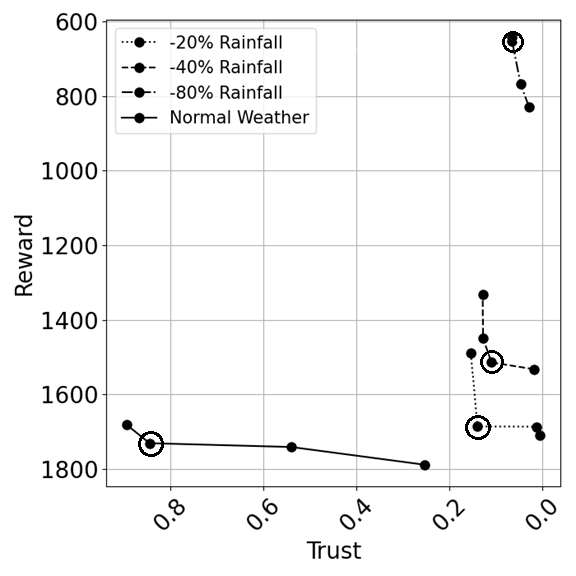}}
\caption{Pareto fronts under scenarios of precipitation reduction. The circled point represents the selected policy.} \label{fig:pft1}
\end{figure}

\begin{table}
\centering
\begin{tabular}{| p{5.4cm} | p{2.4cm} | p{2.6cm} | } 
 \hline
 \diagbox[innerwidth=5.4cm]{Precipitation reduction}{recommendation} & Trust-aware  & Trust-agnostic \\ 
 \hline
 \hline
  $-20\%$ &   &      
 \\ 
 \hline
 Yield (kg/ha) & 8924 & 8930   
 \\
  \hline
 Nitrogen input (kg/ha) & 190 & 160   
 \\
  \hline
 Nitrate leaching (kg/ha) & 0.008 & 0.008   
  \\
  \hline
  Fertilization frequency & 3 &  5  
 \\
  \hline
 Optimal window fertilizer count & 3 &  3 
  \\
  \hline
  Total Reward & 1678 & 1724 
  \\
  \hline
  Trust Score & 0.14 & 0.001 
 \\
  \hline  \hline
   $-40\%$ &   &      
 \\ 
 \hline
 Yield (kg/ha) & 8225 & 7928 
 \\
  \hline
 Nitrogen input (kg/ha) & 190 & 140 
 \\
  \hline
 Nitrate leaching (kg/ha) & 0.0006 & 0.0006 
  \\
  \hline
  Fertilization frequency & 2 &  6  
 \\
  \hline
 Optimal window fertilizer count & 2 &  3 
  \\
  \hline
  Total Reward & 1524 & 1534 
  \\
  \hline
  Trust Score & 0.11 & 0.0001 
 \\
  \hline \hline
    $-80\%$ &   &     
  \\ 
 \hline
 Yield (kg/ha) & 4216 & 4360 
 \\
  \hline
 Nitrogen input (kg/ha) & 190 &  100 
 \\
  \hline
 Nitrate leaching (kg/ha) & 0.005 & 0.005  
  \\
  \hline
  Fertilization frequency & 2 & 5   
 \\
  \hline
 Optimal window fertilizer count & 2 & 3  
  \\
  \hline
  Total Reward & 642 & 809 
  \\
  \hline
  Trust Score & 0.067 & $1.3\times10^{-6}$ 
 \\
  \hline
\end{tabular}
\caption{Comparison of different policies when precipitation decreases}\label{table:rainImpact}
\end{table}

Figure \ref{fig:pft1} illustrates the Pareto fronts generated under scenarios of reduced precipitation. Compared to the scenario of increased temperature, where the trust model demonstrated robustness under small increases, precipitation reductions considerably constrain the feasible region, limiting achievable trust scores. Even moderate rainfall decreases (-20\% ) lead to substantially contracted Pareto fronts, highlighting pronounced trade-offs between reward and trust.

Under moderate precipitation reductions (-20\%), the trust-aware policy achieves comparable yields to the normal weather scenario but experiences notably diminished trust scores. As detailed in Table \ref{table:rainImpact}, this decline in trust primarily arises from significantly lower nitrate leaching due to reduced rainfall, negatively impacting the benevolence of trust. When precipitation deficits intensify (-40\%), yields decrease noticeably, further lowering the trust scores for the trust-aware recommendations.

In extreme drought scenarios (-80\%), both trust and reward metrics sharply decline. Despite trust-aware policies continuing to offer similar fertilizer applications and managing complexity better than trust-agnostic ones, the drastic reduction in yield and minimal nitrate leaching severely impair the trust model's applicability and farmers' perception of trustworthiness becomes severely compromised

Table \ref{table:rainImpact} further elucidates practical implications: trust-aware policies consistently demand more fertilizer applications than trust-agnostic approaches across all precipitation reduction scenarios. However, the corresponding increase in nitrogen input coupled with significantly reduced nitrate leaching results in diminished reward and trust scores. These findings highlight the necessity for recalibrating or adapting the environmental aspects of the trust model to better reflect farmers' preferences under reduced rainfall scenarios.

\section{Conclusion and outlook}

In this work, we introduced a mathematical trust model and integrated it into an MORL framework to generate optimal agricultural management policies on a crop simulator. Our primary aim was to maximize agricultural productivity while explicitly incorporating farmers' trust considerations. We conducted a detailed survey involving 71 farmers, primarily from Iowa, resulting in 54 high-quality, usable responses. Using these responses, we developed and validated a quantitative trust model based on the three-dimensional trust model of ability, benevolence, and integrity. This model was then incorporated directly into the reinforcement learning training process, enabling real-time trust feedback alongside traditional performance metrics such as crop yield, nitrogen use, and nitrate leaching.

Simulation results demonstrated that the generated trust-aware optimal policies effectively balanced agricultural performance and farmer trust compared to expert-derived and trust-agnostic policies. In scenarios involving climate variability, although trust-aware policies achieved slightly lower total rewards than their trust-agnostic counterparts, they consistently resulted in significantly higher trust scores. Specifically, under moderate temperature increases (up to +2°C), the trust model exhibits relative robustness, supporting policies that sustained both trust and yield.

However, under conditions of reduced precipitation — even at modest levels — the trust model’s effectiveness declined more rapidly, with sharply reduced trust scores and a constrained policy space. In extreme climate conditions, such as +5°C temperature rise or an 80\% rainfall reduction, the model’s performance was notably limited. These findings indicate constraints in the current trust model’s adaptability to severe climate variability, likely due to its development being based on farmer survey data collected under the assumption of normal weather patterns. 

To address this limitation, future work should focus on updating survey to include a larger and more diverse farmer participant - targeting over 200 respondents - and enhancing model robustness by explicitly incorporating climate variability. Consequently, integrating dynamic and adaptive trust mechanisms into AI systems that can respond to evolving climate conditions and shifting farmer preferences will be essential. Our trust-centric MORL framework also holds significant potential for application in other domains where user trust and acceptance are critical, such as personalized healthcare and autonomous transportation. By prioritizing trust in AI development, we pave the way for solutions more closely aligned with human values, expectations, and practical needs.

\section{Ethical Approval and Informed Consent Statement}
All experimental procedures involving human participants were conducted in accordance with the relevant institutional and national guidelines and regulations. Ethical approval for the study was obtained from the Institutional Review Board (IRB) of the University of Iowa under approval reference number 202405307, dated 05/22/2024.

All participants provided written informed consent prior to their inclusion in the study. The privacy rights of all participants have been fully observed, and all data were anonymized to maintain confidentiality.

\section{Acknowledgements}

This material is based upon work supported by the National Science Foundation under grant numbers 2226936 and 2420405 and the U.S. Department of Education under grant number ED\#P116S210005. Any opinions, findings, and conclusions, or recommendations expressed in this material are those of the authors and do not necessarily reflect the views of the National Science Foundation and the U.S. Department of Education.

\section{Author contributions: CRediT}

\textbf{Zhaoan Wang}: Conceptualization, Data curation, Methodology, Software, Validation, Visualization, Writing - Original Draft, Writing - Review and Editing. \textbf{Wonseok Jang}: Conceptualization, Data curation, Methodology, Formal analysis, Writing - Review and Editing. \textbf{Bowen Ruan}: Conceptualization, Writing - Review and Editing, Funding acquisition. \textbf{Jun Wang}: Conceptualization, Writing - Review and Editing, Project administration, Funding acquisition. \textbf{Shaoping Xiao}: Conceptualization, Methodology, Writing - Original Draft, Writing - Review and Editing, Project administration, Funding acquisition.

\end{document}